\documentclass[sigconf,nonacm]{aamas}

\usepackage{balance} 

\usepackage{todonotes}
\usepackage{amsmath}
\usepackage{pdfpages}
\usepackage{amssymb}
\usepackage{booktabs}
\usepackage{pifont}
\usepackage{tabularx}
\usepackage{ragged2e}
\newcommand\changed[1]{{#1}}

\title{Planning Ahead with RSA: Efficient Signalling in Dynamic Environments by Projecting User Awareness across Future Timesteps}

\author{Anwesha Das}
\affiliation{
  \institution{Saarland University}
  \city{Saarbr\"ucken}
  \country{Germany}}
\email{adas@lst.uni-saarland.de}

\author{John Duff}
\affiliation{
  \institution{UCLA}
  \city{Los Angeles}
  \country{USA}}
\email{duff@ucla.edu}

\author{Jörg Hoffmann}
\affiliation{
  \institution{Saarland University}
  \city{Saarbr\"ucken}
  \country{Germany}}
\email{hoffmann@cs.uni-saarland.de}

\author{Vera Demberg}
\affiliation{
  \institution{Saarland University}
  \city{Saarbr\"ucken}
  \country{Germany}}
\email{vera@lst.uni-saarland.de}

\begin{abstract}
Adaptive agent design offers a way to improve human-AI collaboration on time-sensitive tasks in rapidly changing environments. In such cases, to ensure the human maintains an accurate understanding of critical task elements, an assistive agent must not only identify the highest priority information but also estimate how and when this information can be communicated most effectively, given that human attention represents a zero-sum cognitive resource where focus on one message diminishes awareness of other or upcoming information. We introduce a theoretical framework for adaptive signalling which meets these challenges by using principles of rational communication, formalised as Bayesian reference resolution using the Rational Speech Act (RSA) modelling framework, to plan a sequence of messages which optimise timely alignment between user belief and a dynamic environment. The agent adapts message specificity and timing to the particulars of a user and scenario based on projections of how prior-guided interpretation of messages will influence attention to the interface and subsequent belief update, across several timesteps out to a fixed horizon. In a comparison to baseline methods, we show that this effectiveness depends crucially on combining multi-step planning with a realistic model of user awareness. As the first application of RSA for communication in a dynamic environment, and for human-AI interaction in general, we establish theoretical foundations for pragmatic communication in human-agent teams, highlighting how insights from cognitive science can be capitalised to inform the design of assistive agents.
\end{abstract}


\newcommand{\BibTeX}{\rm B\kern-.05em{\sc i\kern-.025em b}\kern-.08em\TeX}

\begin{document}


\pagestyle{fancy}
\fancyhead{}


\maketitle 


\section{Introduction}\label{sec:introduction}



\changed{In high-stakes environments, multiple critical events often unfold simultaneously, making it difficult for human overseers to maintain situational awareness of rapidly evolving scenes. Consider an air traffic controller managing several aircraft during a storm, or a power grid control room, as illustrations. In such settings where human perceptual lapses (due to overload and eroding situational awareness) carry high risk, assistive AI agents can support operators who must multitask under increased cognitive demands to sustain effective oversight \cite{barshi2023people, shappell2006human}.}

\changed{The efficacy of AI assistance, however, is limited by the fact that human attentional resources are finite. Once a message captures and engages attention, it temporarily constrains an individual's ability to perceive and process new information \cite{wickens2008multiple}. In dynamic settings with multiple concurrent events, the timing and content of communications influence whether human partners maintain situational awareness or experience overload \cite{endsley2017toward}. Poorly-calibrated communication has severe consequences in such safety-critical situations (see \citealp{ancker2017effects}, \citealp{gulcsen2025effect} and \citealp{mongan2020artificial} for documented accidents). A pattern follows: although today's intelligent systems are adept at identifying all important (or critical) information quickly amidst vast data streams, they do not yet (sufficiently) take into account human cognition when communicating this information \cite{mussi2025human}.}

\changed{The true value of AI assistance, hence, comes about not from indiscriminately alerting human partners of all critical information but depends instead on the assistive agent's capacity to deliver alerts that are timed and framed to align with individual operators' evolving mental model (beliefs and attentional resources) of the world \cite{gmytrasiewicz2001rational, collins_building_2024}.}

\begin{figure}[t]
\centering
\includegraphics[width=\linewidth]{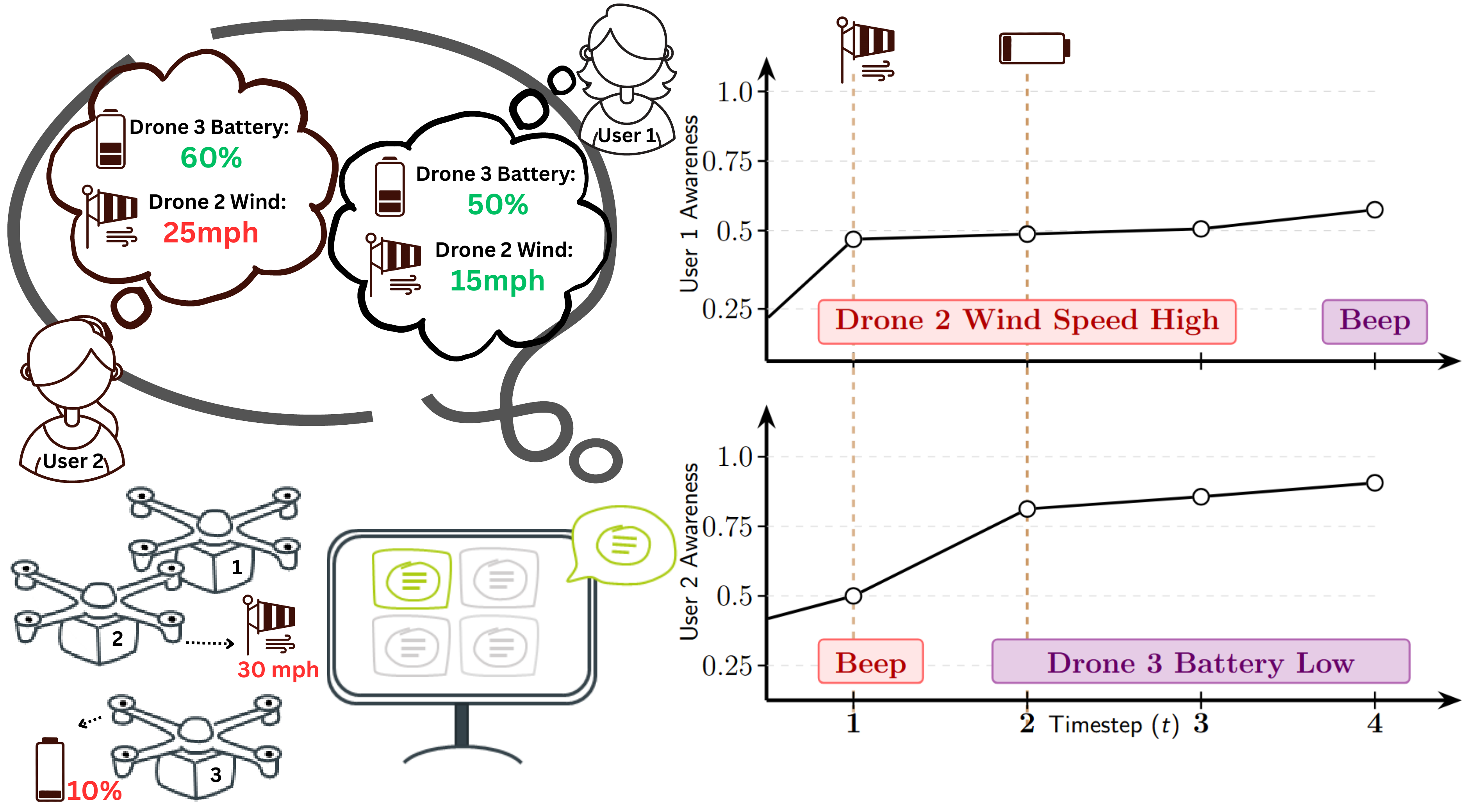}
\caption{Impact of individualised alert sequences (x-axis) for distinct users in the same environment on user awareness (y-axis). Timesteps (x-axis) are annotated with the onset of critical events. Our model uses recursive Bayesian reasoning to construct user-specific models of how each user updates their world beliefs in response to alert sequences, optimising for global situational awareness over time.}
\end{figure}

Suppose you are an operator who must oversee a swarm of fire-fighting drones -- a dynamic, high-stakes environment where time is constrained. When concurrent risks emerge: (a) drone 2 facing high wind-speed conditions, shortly followed by (b) drone 3's imminent battery failure, each requires urgent but distinct alert messages from the AI system. It would be prudent for a good assistive system to understand you (your prior beliefs, attention, and environment) to be able to choose a sequence of alerts which would most quickly lead you to a correct awareness of these risks \cite{collins_building_2024}. For example, in this scenario (shown in Fig. 1), one solution would be to deliver a precise alert about the critically high wind-speed before moving to the battery issue, although this may delay the timing of the latter alert. For some users (e.g., User 1 in Fig. 1), this would be a good choice, as both issues must be dealt with, and attempting a shorter, less specific message about the wind could leave a user confused. However, if you (User 2 in Fig. 1) were already anticipating the high wind-speed, but are unaware of the battery issue, an AI assistant should take advantage of this existing knowledge: it could choose the shorter wind-speed message, knowing it would be correctly interpreted, and thus allow for a more timely notification about the unexpected battery failure.

\changed{Our work develops a theoretical framework for optimising the delivery of alerts (or messages) that takes into account the human partner's attention and awareness when choosing how and when to communicate critical information. } By reasoning about both the world and the human’s evolving mental model of it, we aim to enable assistive AI agents that can generate messages that are truly context and human-aware. As \citet{miehling_language_2024} attest, AI agents built atop cognitive models of rational human communication---capable of reasoning about human partners' beliefs, goals, and attentional limits---can enable uniquely sophisticated partnerships. This capability grows increasingly necessary as the AI we interface with expands out from constrained, screen-based interactions to more complex, high-stakes, and dynamic environments: (semi-)autonomous vehicles navigating complicated urban traffic \cite{lin_decision-oriented_2024}, robotic surgeons adapting to complications \cite{sigari_medical_2021}, or drones coordinating disaster responses \cite{khan_exploratory_2019}.

First, to model rational communication, we draw from research in pragmatics, in particular the Rational Speech Acts Framework (RSA) \cite{goodman2016}, a computational framework which uses Bayesian reasoning to model interactions between speakers and listeners. It captures how speakers tailor messages by recursively simulating the listener’s beliefs about messages they hear, allowing RSA-based approaches to enrich data-driven models on tasks like scene description, referential expression generation and instruction following \cite{monroe2017colors, andreas2016reasoning,fried2017unified}. However, vanilla RSA lacks multi-step reasoning and only optimises for single-turn utterances (or alerts), making it ill-suited for dynamic monitoring tasks that require incremental, adaptive communication. For instance, vanilla RSA myopically optimises current alerts (like User 1's wind-speed alert) without considering their time impact on future communications.
Even though neural models offer alternatives, they produce overly verbose outputs violating human pragmatic preferences -- problematic when communicative efficiency is key \cite{ma2025,tang2024}.

To address this gap, we propose a pragmatic signalling framework that extends RSA with finite-horizon lookahead planning. Our approach preserves RSA’s strengths of recursive communication modelling while extending it to real-world settings through reasoning about sequences of messages over multiple timesteps. Thus, it models (a) the human’s evolving estimated beliefs, (b) the dynamics of the environment, and (c) the projected communicative utility of different alert sequences. By estimating dynamic, user-specific belief updates up to a planning horizon, our system strategically decides \textit{what}, \textit{how}, and \textit{when} to communicate to the human partner, ensuring that messages are conducive and contributive to the operator’s pre-existing as well as future knowledge and attention. \changed{This adaptive, context-sensitive approach, grounded in conversational efficiency, establishes theoretical foundations for assistive agents that can reason about the temporal dynamics of human attention and belief evolution in changing environments.}

\section{Probabilistic Pragmatic Models}

The Rational Speech Act (RSA) framework \cite{frank2012, goodman2016, degen2023} formalises effective communication as Bayesian optimization, where conversational partners reason recursively about each other’s mental states. Pragmatic behaviour, moving beyond literal communication alone, arises naturally from this recursive social inference process. Literal meanings are enriched through strategic inferences from prior knowledge and balance between informativity and cost. Classic RSA models can explain phenomena like scalar implicature (`some' implying `not all') \cite{franke2011, goodman2013} and ambiguous reference resolution \cite{frank2012, monroe2017, hawkins2021} by simulating how listeners infer speaker intentions, and vice versa.
This has allowed RSA to be successfully applied in diverse settings from reference games \citep{frank2012} to grounded visual scene descriptions \citep{cohngordon2018}. However, vanilla RSA's successes are fundamentally limited: the framework is optimised for single-turn messages, useful only in static environments. So, while RSA determines \textit{how} to convey important information in isolation, it cannot plan \textit{what} to prioritize or \textit{when} to intervene as environments evolve---a consequential limitation in dynamic, multi-step settings, like ours.


To address these limitations, previous work has pursued partial extensions toward temporal reasoning. \citet{cohngordon2019incremental} extend RSA to score partial messages based on a projected listener interpretations of probable continuations. Their \textit{incremental RSA} model prioritizes packing information as early in the message as possible, but applies this only to the next timestep utility of these partial prefixes---appropriate perhaps for modelling human behaviour, but insufficient for global sequence efficiency.

To achieve our goals, we develop an alternative RSA extension, which also prioritizes temporal efficiency within a multi-timestep sequence, but does so by considering finite-horizon projections of future message sequences. To estimate projected sequence efficacy, it is necessary to project how incremental interpretation affects user priors at future timesteps. This dynamic prior accumulation has been previously considered within RSA for multi-turn conversations \citep{anderson2021}. We extend it here for a practical setting, and show how dynamic representation of user belief can be deployed within RSA as part of multi-step planning in an environment which is itself changing over time.

\section{Problem Formulation}
\label{sec:formal}

We formalise the setting described above as a pragmatic assistive signalling framework, where an AI system (Speaker, $S_2$) communicates using utterances (messages) with the human operator (Listener, $L_1$) to help them track the state of the world over multiple timesteps (finite horizon $ 1 \le t \le H$). This interaction is described by the tuple $(S, \mathcal{P}, U, H, B, R)$, where $S$ is the set of true world states, $\mathcal{P}$ the set of properties constituting $S$, $U$ a space of utterances the system can use to signal at time $t$, $H$ the planning horizon, i.e., the maximum number of timesteps considered when selecting utterance sequences, $B$ the distribution of the human operator's beliefs, and $R$ the reward function.

The world state $S$ evolves dynamically, with each timestep $t$ yielding $s_t: \mathcal{P} \mapsto V$ that assigns values to all properties $p \in \mathcal{P}$ (e.g., $s_t(\text{drone~4 battery level}) = 80\%$). While the system $S_2$ has full access to the true world state, the human $L_1$, however, cannot directly observe the complete $s_t$. Instead, their beliefs about the world $b_t$ are a noisy and lagging estimate of likely states, following changes in $S$ only through observations of individual $p$. Thus, due to distractions or perceptual overload, $L_1$ often has outdated, false beliefs. This results in low situational awareness, which can be risky, especially if there are critical properties $Cr_t(p) = 1$. Certain properties become critical when $s_t(p)$ takes a value in a predefined critical (risk) region $StatCr(p) \subseteq V$,
triggering $Cr_t(p) = 1 $ (e.g., $StatCr(\text{drone~4 battery level}) =[0, 15\%]$ or $StatCr($drone~4 rotor condition$) = [${\textit{off}}]). \changed{Formally, $Cr_t(p) = \mathbf{1}[s_t(p) \in StatCr(p)]$.}

To help $L_1$ sustain high situational awareness and to communicate risks, $S_2$ selects utterances $u \in U$ from a predefined, finite lexicon $\mathcal{M}$ where each $u$ maps to a subset of $P$ via $\mathcal{M}: U \mapsto 2^\mathcal{P}, \; \mathcal{M} \in \{0, 1\}^{|U| \times |\mathcal{P}|}$. To illustrate, consider a simple lexicon $\mathcal{M}$ consisting $3$ utterances of various specificities: \textit{``Drone 4 battery level is low at 10\%"}, \textit{``Battery Low"}, and \textit{``Beep"}, all of which might direct attention to $Cr_t(p = \text{drone 4 battery level})=1$. Ergo, matrix $\mathcal{M}$ tracks the $(u,p)$ mapping permitting variations in both informativeness and length, when referring to a property $p$.

The assistive system $S_2$’s objective is to generate a sequence of utterances $(u_1 , \dots , u_H)$ that maximises reward $r_t$. The reward function $R$ incorporates the $L_1$'s estimated belief distributions $B_t$ to calculate the effectiveness of each utterance sequence in maximising belief alignment ($b_t \approx s_t$). The $R$ can be formulated based on task or domain-specific goals---for our purposes, the goal is sustaining situational awareness about critical properties or risks.


\section{Models}
Let us now demonstrate how we can approach this problem using and extending RSA. We start with the Vanilla RSA agent and incrementally build in (i) belief updates, (ii) user-specific priors, and (iii) finite-horizon planning to help accomplish the task of pragmatic assistive signalling in dynamic environments.

\subsection{Vanilla RSA: Modelling Pragmatic Reasoning}

A \textbf{Literal Speaker $S_0$} is one which stochastically selects utterances based solely on whether $u$ literally means (or maps to) property $p$, without modelling the listener's beliefs or informational needs. This is a naive interface. Using lexicon $\mathcal{M}$, the conditional probability $S_0(u \mid p)$ is split equally between every $u$ whose meaning contains the target $p$:

\begin{equation}
	S{_0}(u \mid p) \propto \frac{
		\exp\left( \log P(p\mid u)\right)
	}{
		\sum_{u'} \exp\left( \log P(p \mid u')\right)
	}
\label{eq:vanilla_S0}
\end{equation}
\changed{Here, $P(p \mid u)$ is derived from the lexicon semantics: $P(p \mid u) \propto \mathbf{1}[\mathcal{M}(u,p)=1]$, yielding a uniform distribution over utterances that refer to $p$.}

At a level higher, the \textbf{Pragmatic Listener $L_1$}, reasons about the speaker’s communicative intent. Given utterance $u$, $L_1$ infers which property $p$ a $S_0$ speaker would most likely have intended to refer to, based on the speaker’s distribution $S_0(u \mid p)$ and their own prior belief $P(p) = \frac{1}{|\mathcal{P}|}, \; \forall p \in \mathcal{P}$:
\begin{equation}
	L{_1}(p \mid u) = \frac{
		S_0(u \mid p) \cdot P(p)
	}{
		\sum_{p'} S_0(u \mid p') \cdot P(p')
	}
\label{eq:vanilla_L1}
\end{equation}

The top layer, \textbf{Pragmatic Speaker $S_2$}, extends this nested reasoning to select utterances that most effectively convey the intended property $p$ to $L_1$. $S_2$ aims to minimise the listener’s uncertainty (or surprisal) about the intended meaning: $-\log L_1(p \mid u)$, adjusted by utterance costs (e.g., length, or complexity). Even though RSA allows an unbounded hierarchy of nested agents ($S_n(L_{n-1})$), studies suggest that a depth of two sufficiently captures the benefits of pragmatic reasoning in most communicative settings \citep{yuan2018understanding}.

Nonetheless, vanilla RSA presents two fundamental limitations that make it insufficient for efficient signalling in dynamic environments: (1) it operates on a single utterance in isolation, lacking explicit models of how listener beliefs evolve and propagate across multiple steps; and (2) it is limited to the selection of an optimal message given a target property $p$ as the goal. That is, while vanilla RSA can reason about a listener's interpretation of utterances $u \in U$ at current time $t$, it cannot track how each $u$ revises their beliefs and priors for subsequent timesteps. And because it provides no principled mechanism for determining which $p \in P$ merits the listener's attention, especially when multiple $p$ may be simultaneously critical, the decision of \textit{what} to communicate remains unresolved. One could repetitively invoke RSA at each timestep, or one could apply heuristic property selection, but such approaches fundamentally sidestep the core challenge: in dynamic environments where context and relevance evolve rapidly, intelligent assistive systems must not only utter a maximally pragmatic signal for current $t$ but also deliberately determine which property of the evolving system deserves attention, and when.

\subsection{Baseline Model: Tracking Beliefs over Time}


We address vanilla RSA's limitations by introducing temporal belief updates and reward-based property selection to modify the pragmatic speaker $S_2$. The $S_0$ and $L_1$ agents remain as defined in Equations \ref{eq:vanilla_S0} and \ref{eq:vanilla_L1}, but now operate with the assumption that the system has access to previous beliefs $b_{t-1}$ and world context $S$ (see Problem Formulation).

\subsubsection{Belief Tracking over Time.} 
The motivation for extending RSA lies in its Bayesian structure, which naturally supports adaptive approaches. In Bayesian models, the posterior from one timestep becomes the prior for the next -- `\textit{today's posterior is tomorrow's prior}' \cite{lindley1972bayesian} -- lending itself well to tracking how user beliefs evolve in response to sequences of utterances, requiring temporal indexing across timesteps.

We operationalise how a user's attention evolves after an alert at current time $t$ given the updated $L_1$ distribution. Function $AT$ maps the $L_1$ to an updated attention distribution as $AT: (L_1, at) \mapsto at$, where, for each property $p \in \mathcal{P}$, the updated attention $at_t(p)$ reflects the likelihood of the user attending to or understanding property $p$ after observing utterance $u_t$. Formally, $\forall p \in \mathcal{P}, \forall t > 0$:

\begin{equation}
at_t(p_t \mid u_t) := AT[L_1(p_t \mid u_t; b_{t-1})]
\label{eq:attention}
\end{equation}

Next, function $B$ estimates how a user's belief state $b_t$ evolves given the prior belief $b_{t-1}$, current system state $s_t$, and attention $at_t$.\footnote{Since real user beliefs are not directly observable, $B$ serves as a computational approximation. More details about alternate variants of functions $(AT, B, R)$ are in the Supplementary Materials.} For each property $p \in \mathcal{P}$ and value $v \in \mathcal{V}$, the updated belief $(\forall t > 0)$ is:

\begin{equation}
\begin{aligned}
& \hspace{-0.3cm} b_t(u_t \mid p_t) = B(b_{t-1}, s_t, at_t)(p)(v) := \\
& \hspace{0.5cm}
at_t(p) \cdot P(v = s_t(p)) + (1 - at_t(p)) \cdot b_{t-1}(p)(v)
\end{aligned}
\label{eq:beliefs}
\end{equation}

where $P(v = s_t(p)) = 1$ if $v$ is the true value of $p$ at $t$, and $0$ otherwise. High attention ($at_t(p) \to 1$) leads to stronger alignment with the true system value $s_t(p)$, while low attention ($at_t(p) \to 0$) preserves the prior belief. This formulation allows the model to track how each potential utterance $u \in U$ would influence user beliefs.

\subsubsection{Rewards.} To determine which property deserves attention based on overall situational awareness and task goals -- we define a reward function that quantifies the communicative value of each utterance $u_t$ in terms of its effect on belief alignment $b_t$ with the true system state $s_t$, weighted by the task-relevance of each $p$ -- here, whether $p_t$ is critical:
\begin{equation}
R_t(u_t \mid s_t, Cr_t) = \sum_{p \in \mathcal{P}} b_t^{u_t}(p)(s_t(p)) \cdot Cr_t(p), \; \; \forall t>0 \\
\label{eq:reward}
\end{equation}
Intuitively, the reward is a scalar value that measures how well a $u_t$ guides the listener's mental model toward an accurate understanding of currently critical properties ($Cr_t(p) = 1$).

\subsubsection{Pragmatic Speaker.} The baseline $S_2$ now can make more informed $u$ selections which account for both belief evolution and property prioritisation. The optimal utterance at each time $t>0$ is then chosen as:

\begin{equation}
S_2^{\text{baseline}}(u_t \mid s_t, Cr_t) = \arg\max_{u \in U} R_t(u \mid s_t, Cr_t)
\label{eq:baseline_u}
\end{equation}

\changed{The baseline model represents our closest analogue to prior work, now extending vanilla RSA to dynamic environments by building in temporal belief tracking. While it can reason about evolving beliefs, it makes simplifying assumptions: uniform user priors and greedy (single-timestep) temporal reasoning. As such, it serves as the foundation for evaluating more sophisticated variants that we introduce next. We call it \textbf{Dynamic RSA (d-RSA)} for brevity.}

\subsection{Pragmatic Human-Aware Planning}

Having established the baseline model, we introduce two further modifications that improve its adaptive pragmatic signalling ability: user priors and temporal planning. We establish these reasoning capabilities independently before combining them together to develop the full model.

\subsubsection{Reasoning about Users:} The baseline model's uniform priors $P(p) = \frac{1}{|\mathcal{P}|}$ reflect a generic user assumption, treating all listeners as identical entities. Yet real users are diverse -- bringing varied beliefs, biases and contextual awareness that influence how they interpret communicative signals significantly. We capture this heterogeneity by substituting these uniform priors with user-specific priors from the beginning: conditioning the $L_1$ listener on their previous belief state \(b_{t-1}\) and task-relevant contextual knowledge: $P(p \mid b_{t-1}, StatCr)$.

\begin{equation}
\begin{aligned}
& \hspace{-0.5cm} L_1(p \mid u, b_{t-1}, StatCr) \hspace{0.3cm} = \hspace{0.1cm} \\
& \hspace{0.7cm} \frac{
S_0(u \mid p) \cdot P(p \mid b_{t-1}, StatCr)
}{
\sum_{p'} S_0(u \mid p') \cdot P(p' \mid b_{t-1}, StatCr)
}
\end{aligned}
\label{eq:L1_myopic}
\end{equation}

These user-specific priors empower the model to reason about distinct user profiles (noisy users), who may, for e.g., maintain high prior expectations for familiar system properties (e.g., battery), or exhibit attention biases toward aspects they perceive as safety-critical based on prior knowledge (e.g., windspeed).

\subsubsection{Reasoning about Time:} The myopic approaches we've presented above---whether baseline or user-specific---optimise each utterance in isolation, sacrificing long-term goals for immediate rewards. To surmount this, we can extend the Pragmatic Speaker $S_2$ with finite-horizon planning. The planner generates all possible utterance sequences $\mathbf{u}_{1:H} = (u_1, u_2, \dots, u_H)$ up to planning horizon $H$, subject to duration constraints. For each complete sequence, it computes the total cumulative reward:
\begin{equation}
\mathcal{R}(\mathbf{u}_{1:H}) = \sum_{t=1}^{H} R_t(u_t \mid s_t, Cr_t)
\label{eq:cumulative_reward}
\end{equation}
\noindent{\textbf{subject to:}} \begin{equation*}
    \textit{dur}(u_t) = k_{u_t} > 1 \implies u_{t+1}, \dots, u_{t+k-1} \text{ are } \textit{`(X)'}
\end{equation*}

This constraint encodes temporal opportunity costs -- longer utterances lock the system into a multi-step delivery, during which only \textit{`(X)'} messages are legal. These \textit{`(X)'} utterances assign zero attention to all $p$, representing cases where a listener's attentional resources are already committed to process an ongoing multi-step utterance $(\textit{duration}(u_t) = k(u_t) > 1$ timestep). They yield no belief updates and thus low or zero rewards, while also delaying other reward-accruing utterances. This double penalty allows the planner to simulate trade-offs between informativeness and timeliness.

Applied with only uniform priors, the result of this user-agnostic planning would likely result in a less sensitive model. We use this deliberate ablation to isolate the contribution of planning (reasoning about time), \changed{terming it the \textbf{d-RSA + Planning} model.} We term the model with informed priors (Eq. \ref{eq:L1_myopic}) but without planning the \changed{\textbf{d-RSA + Priors} model}, to assess the impact of user-modelling (reasoning about users) individually.

\subsubsection{Reasoning about Users and Time:} Finally, combining user‐specific priors (Eq. \ref{eq:L1_myopic}) with finite‐horizon planning lets the agent track evolving user beliefs and evaluate complete utterance sequences. \changed{This yields our full model, \textbf{d-RSA + Priors + Planning}.} The optimal utterance sequence $U^*$ is the one that maximises cumulative reward over all candidate sequences $\mathbf{u}_{1:H}$:
\begin{equation}
U^* = \arg\max_{\mathbf{u}_{1:H}} \sum_{t=1}^{H} R_t(u_t \mid s_t, Cr_t)
\label{eq:final_seq}
\end{equation}
The pragmatic speaker then executes the $t^{th}$ element of $U^*$ at each timestep $\forall t \in \{1, \dots, H\}$ as $S_2^{\text{full}}(u_t) = (U^*)_t$.

\section{Experiments}\label{sec:experiments}

\begin{figure*}[t]
    \centering
    \begin{tabular}{cc} 
        \textbf{(a)} \includegraphics[width=0.355\textwidth, trim={0 0 5cm 0}, clip]{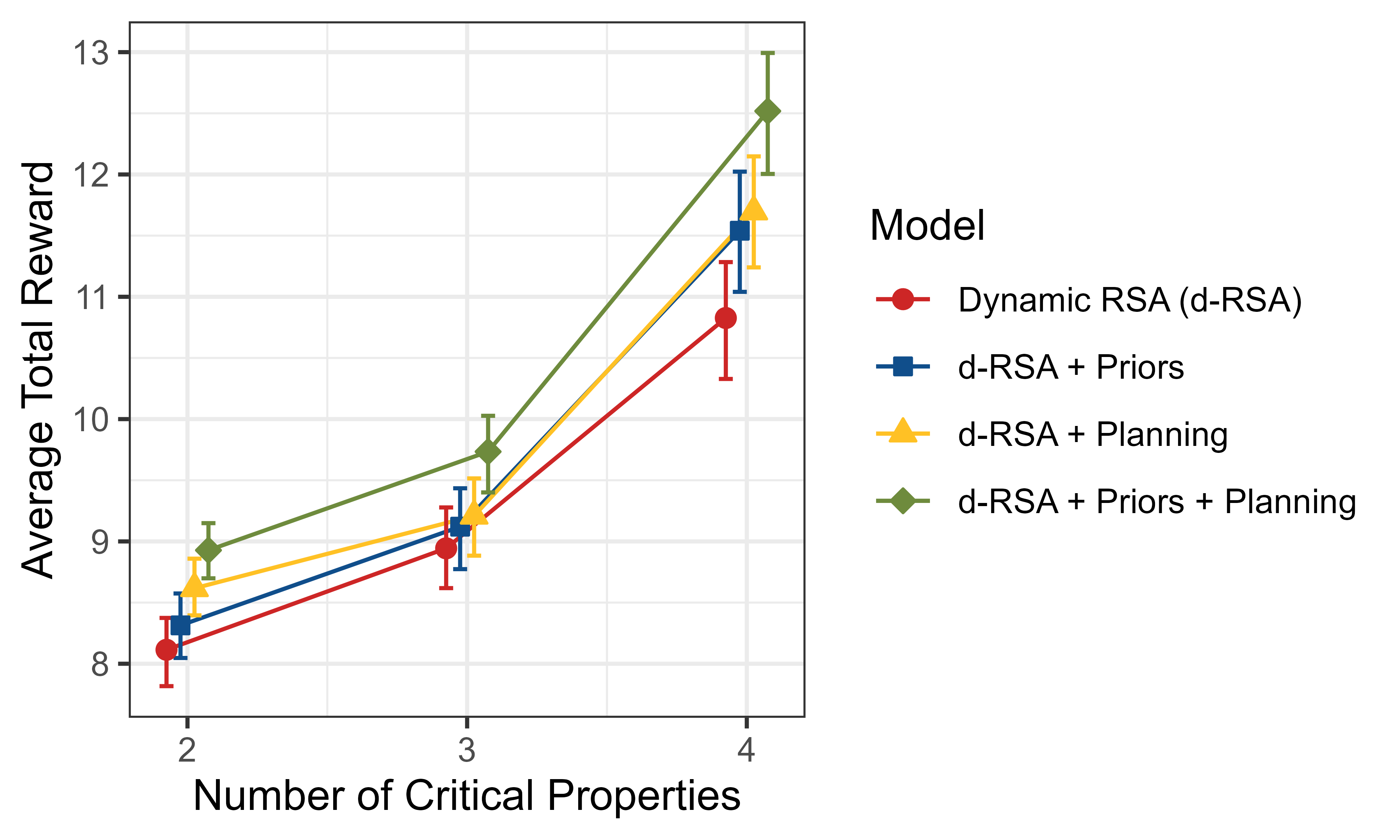}
        &
        \textbf{(b)} \includegraphics[width=0.58\textwidth, clip]{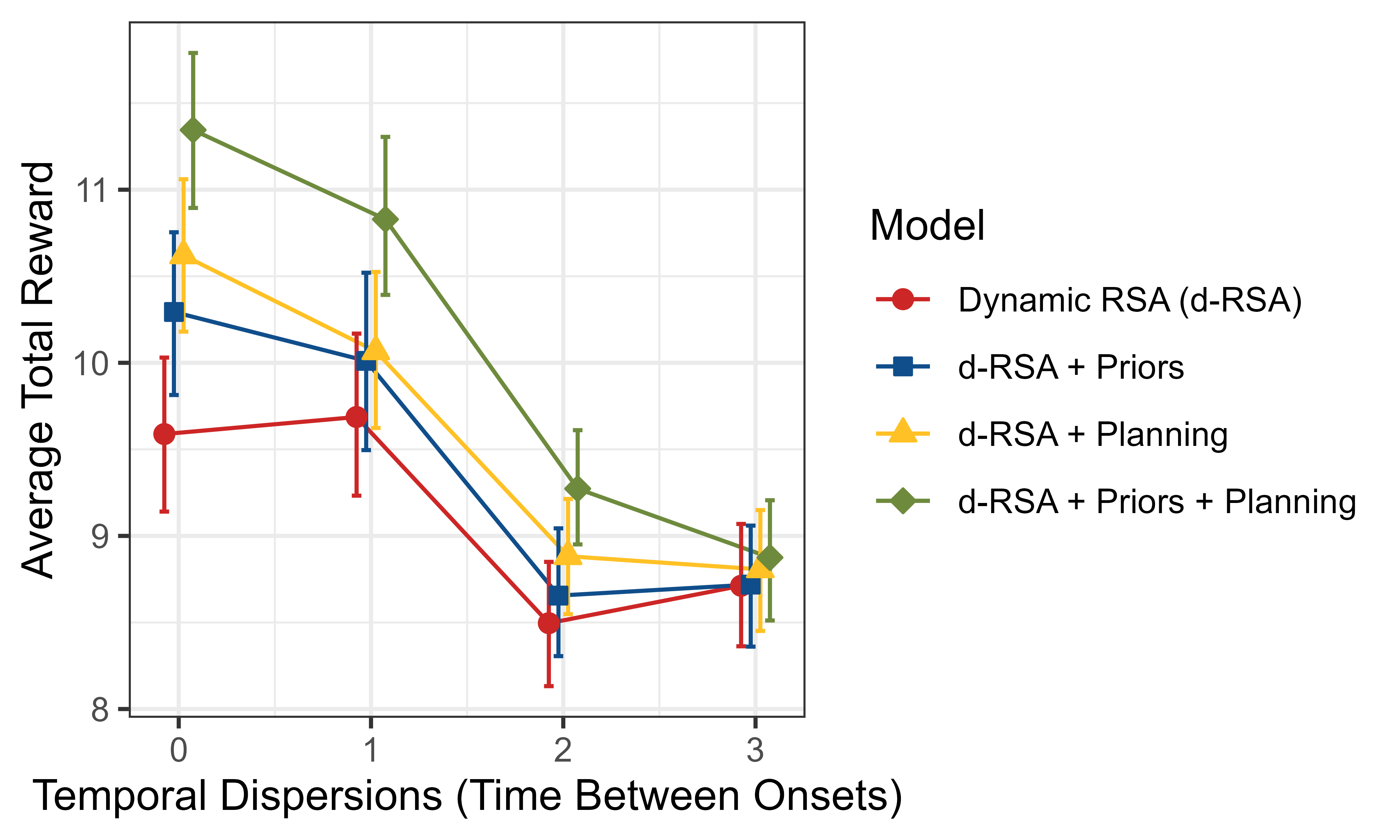}
        \\
    \end{tabular}
    \caption{Total reward (y-axis) for each model variant across scenario factors. (a) Reward as a function of the number of critical properties (higher = more challenging). (b) Reward vs. temporal dispersion of critical onsets (lower = more tightly clustered, i.e., challenging). The full model achieves the greatest relative gains as scenarios become more difficult.}
    \label{fig:2a2b}
\end{figure*}

\subsection{Setup}

\subsubsection{Environment}

To evaluate our model's adaptability across diverse challenging scenarios, we created $800$ trials in a simulated Drone World environment. Each trial involves $4$ drones with $6$ attributes each ($24$ total properties) with their values evolving over $7$ timesteps. Properties are either continuous (e.g. altitude) or binary (e.g. rotor condition). 

Similar to the toy example in Problem Formulation, our lexicon $\mathcal{M}$ contains utterances of various specificities to refer to each property. We define a duration vector $Dur \in \mathbb{N}^{|U|}$, where $Dur_u = k(u)$ denotes the timesteps required to utter $u$. For example, a specific message like `\textit{Drone 4 battery level is low}' consumes 3 timesteps, while a brief `\textit{Beep}' takes only 1. Together, $(\mathcal{M} + k(u))$ defines our lexicon, encoding both $(u, p)$ mapping and temporal cost. We also introduce a special \textit{informationless} utterance \cite{scontras2021practical}: \textit{Silence `...'} which distributes minimal attention uniformly across all $p$ and not only helps capture naturalistic fluctuations of attention (due to scanning, mind-wandering, distractions, etc.) but more importantly provides an alternative to constant alerting when it is not optimal.

We systematically vary trial complexity as follows. \\
\textbf{Critical Properties} (hereafter, $Cr(p)$): Each scenario has $2-4$ $Cr(p)$, sampled randomly from the full set of $24$. Some trials have overlapping $Cr(p)$ features---multiple risks associated with the same drone or attribute, which we quantify with \textit{overlap metric}: the proportion of shared features.\\
\textbf{Temporal Dispersion:} Each $p$ becomes critical at specific timestep $t$. We vary the spacing between them between $0-3$, with the first onset occurring $1 \le t \le 3$.\\
\textbf{User Belief Modelling:} The $L_1$ begins with an initial belief $b_{t=0}(p, v)$, a probability distribution over each property $p$ and its possible values $v$. We simulate diverse (misspecified or noisy) users by varying (i) the accuracy of $b_{t=0}(p, v)$ about critical properties (10\%-80\% correct) and (ii) general situational awareness (low vs. high accuracy and precision for non-critical properties).\footnote{Detailed descriptions in the supplementary material.}

\subsubsection{Implementation}

Our approach employs finite-horizon planning with the objective to maximise (non-discounted) rewards, through a reward designed to integrate RSA with lookahead. For our experiments to fairly evaluate the model design variants, we need to know the optimal plan for each trial and its maximal attainable reward. To this end, we use breadth-first search to enumerate utterance sequences up to horizon $H=7$, enforcing multi-timestep constraints to construct a pruned search tree.

\subsubsection{Analysis} We evaluate the overall performance of the full d-RSA + Priors + Planning model and its two key capabilities:
\begin{itemize}
\item[\textbf{(A)}] \textbf{Multi-step Planning:} Does planning utterance sequences based on their project rewards over a finite horizon, yield better communicative outcomes than selecting the optimal option myopically at every free timestep?
\item[\textbf{(B)}] \textbf{Belief Tracking:} Does selecting utterance sequences by reasoning about the users' estimated beliefs improve signalling compared to assuming generic, uniform priors?
\end{itemize}

\changed{To recall, our analysis compares four models: the baseline Dynamic RSA (d-RSA), which extends vanilla-RSA to a dynamic environment but is greedy and user-agnostic; d-RSA + Priors, which adds user-modelling but remains greedy; d-RSA + Planning, which adds planning but uses uniform priors; and the full d-RSA + Priors + Planning model.}

Comparisons between (d-RSA + Priors + Planning) vs. (d-RSA + Priors) isolate (A), while (d-RSA + Priors + Planning) vs. (d-RSA + Planning) isolate (B).

\section{Results}

\subsection{When is pragmatic planning most useful?}

To understand our full model's strengths, we compare achieved rewards across all $800$ trials against the three model variants: d-RSA, d-RSA + Priors, and d-RSA + Planning. In particular, to extract generalisations about scenarios where our full d-RSA + Priors + Planning model produces the highest gain over each baseline, we fit a linear mixed-effect regression predicting total reward as a function of model, scenario, and user properties, and their interactions, with random intercepts for uncontrolled variance. We apply this same regression approach to all metrics reported in the following sections.

Pairwise comparisons show the d-RSA + Priors + Planning model outperforms all baselines ($p\,$s $< 0.01$), with increases associated with both planning and belief-tracking ($p\,$s $< 0.01$). These elements are particularly valuable in combination: planning benefits most when user-specific beliefs are available ($p= 0.02$). Benefits are largest in the highest-difficulty scenarios with many critical properties or critical properties emerging closer in time ($p\,$s $< 0.01$; see Figures 2a, 2b).

Unsurprisingly, we observe that modelling user beliefs delivers the highest gain when users begin with a higher awareness of critical properties ($p = 0.03$)---compare the scenario in Fig. 1, where user-based planning increases communicative speed by capitalising on partial pre-awareness. Table \ref{tab:examples} presents cases where the model produces a higher value. In scenario (A), four properties become critical in short succession, beginning with two which the user anticipates. The simpler d-RSA and d-RSA + Planning models use long, specific utterances covering only two problems, failing to prioritise those that the user is unaware of. The d-RSA + Priors + Planning model packs in more messages by exploiting user knowledge for shorter alerts. Another strategy that emerges in crowded scenarios for the full model use of anticipatory alerts, raising awareness before criticality -- in (B), where the d-RSA model waits silently before several properties become critical together, the d-RSA + Priors + Planning model uses specific messages directing attention to near-critical values.
This still leaves space for a specific alert once criticality hits, while shifting the users' priors for unanticipated problems, raising the chance that a quick \textit{beep} is informative.



\begin{table*}
\small
\setlength{\tabcolsep}{3pt}
\caption{Example test scenarios, in terms of critical property identities and onsets, and the sequences generated by our models.}
\begin{tabularx}{\textwidth}{@{}l X l l@{}}
\toprule
    & Critical Properties with Onset Time (\textbf{User Unaware}) & Top Sequence [$U^*_{\text{B}}=$ Baseline (d-RSA), $U^*_{\text{F}}=$ Full (d-RSA++)] & $\sum r_t \uparrow$\\
\midrule
    (A) & (D2\_Dist, 2), (D3\_Alt, 3), \textbf{(D3\_Dist, 4)}, \textbf{(D4\_NoFly, 5)} & $U^{*}_{B} = $ {[}`...', `D2 Distance', (X), (X), `D3 Altitude', (X), (X){]} & 9.65\\
    & & $U^{*}_{F} = $ {[}`...', `Beep', `Altitude', (X), `D3 Distance', (X), (X){]} & 12.29 \\
    \midrule
    (B) & (D1\_Wind, 3), (D1\_Dist, 3), \textbf{(D3\_NoFly, 3)}, \textbf{(D4\_Alt, 3)} & $U^{*}_{B} = $ {[}`...', `...', `D1 Wind Speed', (X), (X), `Distance', (X){]} & 5.29 \\
    & & $U^{*}_{F} = $ {[}'D4 Altitude', (X), (X), `D3 No Fly Zone', (X), (X), `Beep'{]} & 11.24 \\
    \midrule
    (C) & (D1\_Batt, 1), \textbf{(D2\_Rotor, 2)}, (D2\_Alt, 3), (D2\_NoFly, 4) & $U^{*}_{B} = $ {[}'D1 Battery', (X), (X), `D2 Altitude', (X), (X), `Beep'{]} & 13.74 \\
    & & $U^{*}_{F} = $ {[}'Battery', (X), `D2 Rotor', (X), (X), `Altitude', (X){]} &  19.08 \\
    \midrule
    (D) & (D1\_Batt, 1), (D1\_Wind, 1), \textbf{(D3\_Alt, 1)} & $U^{*}_{B} = $ {[}'D1 Battery', (X), (X), `D1 Windspeed', (X), (X), `Beep'{]} & 8.98\\
    & & $U^{*}_{F} = $ {[}'D3 Altitude', (X), (X), `D1 Battery', (X), (X), `...'{]} & 13.47 \\
\bottomrule
\end{tabularx}
\label{tab:examples}
\end{table*}

\subsection{`How' is a message spoken?}

We also investigate message choices by examining alert behaviour differences from the baseline. Message-type entropy diagnoses variability across specificity levels in each model's highest-ranked sequence. The d-RSA + Priors + Planning model produces the highest entropy ($p$s $< 0.01$), particularly supported by planning ($p< 0.01$), while  d-RSA and d-RSA + Priors models prefer fully-specific alerts ($p$s $< 0.01$). Belief-tracking increases entropy specifically in easier scenarios with fewer, more spread-out critical properties and for users who begin with better awareness ($p$s $< 0.01$).

Entropy is largely being driven by an increased adoption of less specific messages in the full model. Measuring message specificity as the ratio of unambiguous (two-feature) to ambiguous (single-feature alerts and beeps) messages, this ratio is lowest in the full model ($p$s $< 0.01$), driven downward by both planning and belief-tracking, and especially by their combination ($p$s $< 0.01$). Both elements produce the largest reductions in specificity when users begin with good awareness ($p$s $< 0.01$; see Figure 3a). Planning furthermore reduces specificity when critical properties are temporally clustered ($p< 0.01$), when efficiency has the greatest value. See scenario (C), where the full d-RSA + Priors + Planning model uses less specific utterances for two better-known properties, reserving time to describe unexpected problems in detail.
That is, when the model knows a user begins with priors that will lead them to resolve less specific messages correctly, it takes advantage of that knowledge, making it more likely to use non-specific messages.

In sum, our full model improves on less sophisticated systems by mixing complete descriptions with carefully-selected, less-specific alerts based on user knowledge and message density, using non-specific alerts when they would be correctly interpreted, where possible.

\begin{figure*}[t]
    \centering
    \begin{tabular}{cc} 
        \textbf{(a)} \includegraphics[width=0.42\textwidth, trim={0 0 5cm 0}, clip]{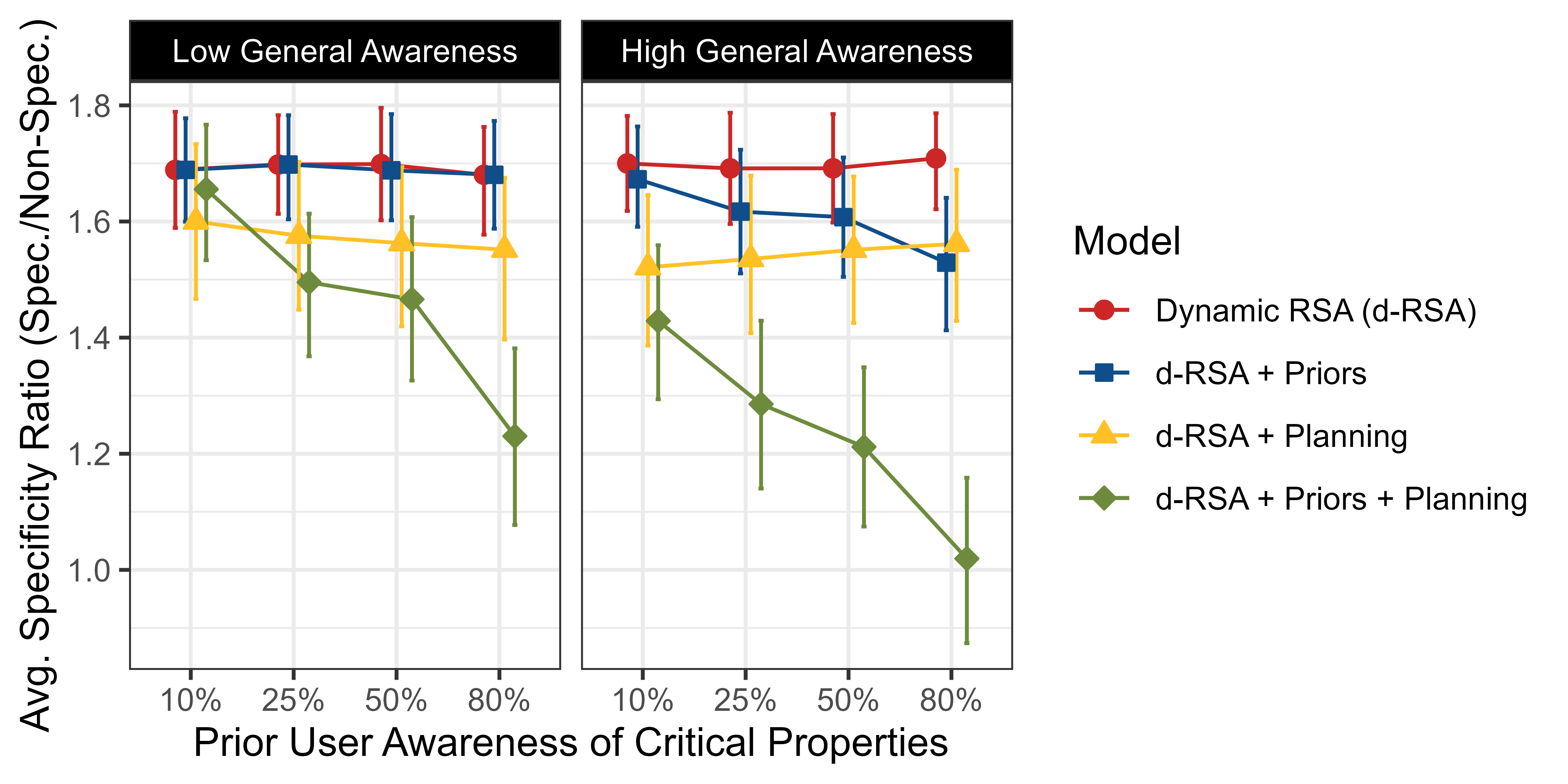}
        &
        \textbf{(b)} \includegraphics[width=0.52\textwidth, clip]{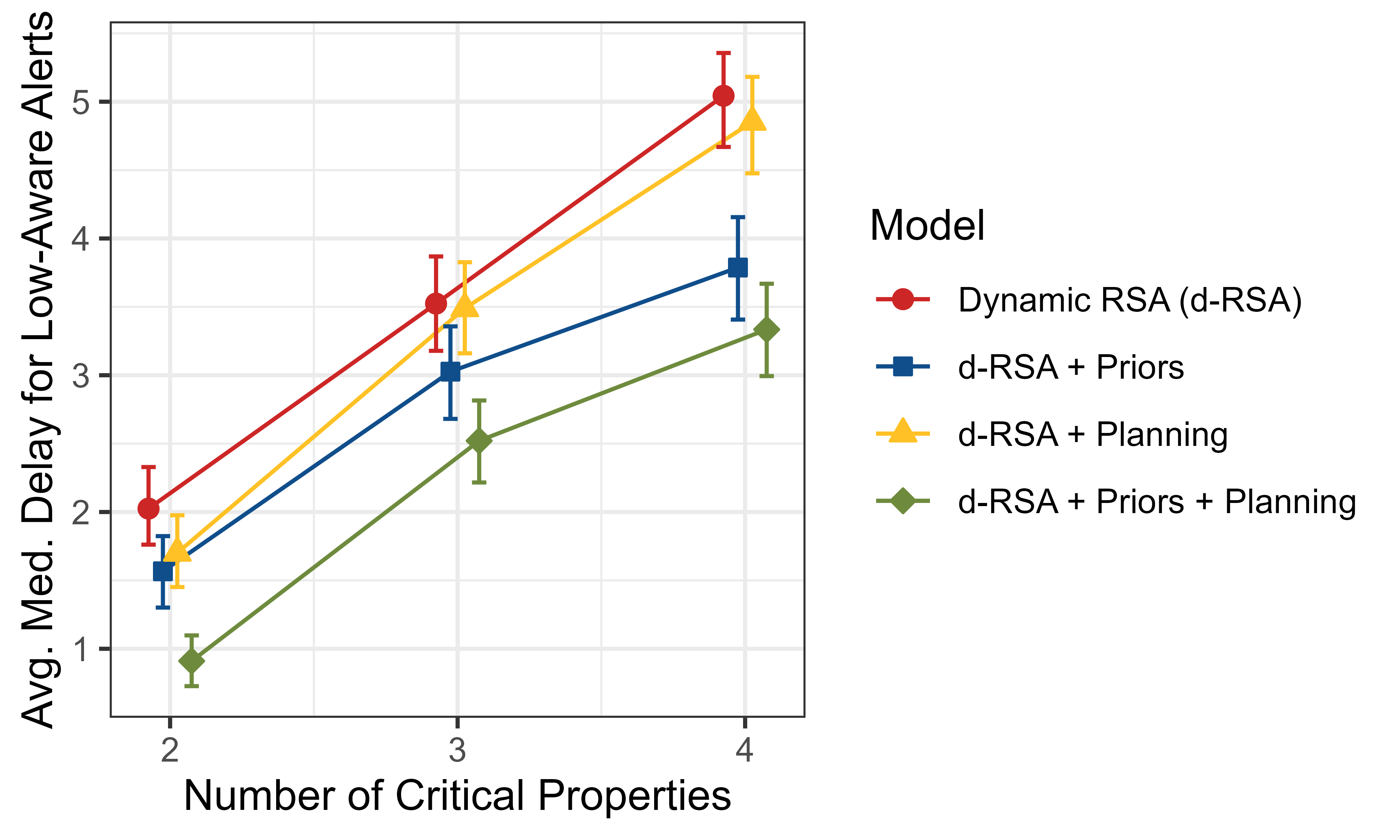}
        \\
    \end{tabular}
    \caption{(a) Message Specificity Ratio (y-axis) vs. Prior User Awareness of critical properties (x-axis). Specificity decreases with higher prior awareness only in the full model, showing its adaptive use of less specific alerts; baseline variants remain less flexible. (b) Median Delay (in timesteps, y-axis) to the First Alert about Low-Awareness Critical Properties vs. Number of Critical Properties (x-axis). The full model is consistently faster as scenario difficulty grows, outperforming baseline variants.}
    \label{fig:3a3b}
\end{figure*}

\subsection{`When' is a message spoken?}

Finally, we investigate how the model times the messages it chooses to send, using the median delay between the onset of low-awareness critical properties and the model's first message expected to shift the distribution of user beliefs towards that property's true value, for each trial.\footnote{Critical properties never mentioned before the horizon are counted as maximum delay of $7$.} The d-RSA + Priors + Planning model achieves shorter delays than all others ($p$s $< 0.01$). Both planning and belief-tracking drive delay down ($p$s $< 0.01$), especially in combination ($p = 0.01$), with belief-tracking bringing the d-RSA + Priors model to second place ($p$s $< 0.01$), outranking the d-RSA + Planning and d-RSA models ($p$s $< 0.01$). Stronger belief-tracking improvements occur in particularly difficult cases with many critical properties ($p = 0.01$; see Figure 3b), dense temporal packing ($p< 0.01$), and higher initial user awareness ($p< 0.01$); see scenario (C).

Substantial delays indicate some risk of delayed response to critical situations. To check how this risk is being managed, we compare this against the median delay for high-awareness critical properties, taking their difference as a proportion of total delay across both (low vs. high awareness) categories. As discussed at the beginning (Figure 1), an adaptive agent should sometimes risk delays for likely-conscious problems to prioritise unconscious ones. Only our full d-RSA + Priors + Planning model significantly prioritises low-awareness properties $(95\% \; CI: 0.02, 0.13)$, as per the difference metric. Belief-tracking is the major contributor ($p< 0.01$), with the d-RSA + Priors model performing well but not significantly favouring low-awareness properties $(95\% \; CI: -0.04, 0.06)$ nor differing from the d-RSA + Priors + Planning model ($p = 0.12$). See scenario (D), where user prior awareness allows our d-RSA + Priors + Planning model to accurately prioritise among three simultaneous problems.


To summarise, our full model improves the timing of the most important alerts by selectively delaying mention of some problems, often prioritising those which the user is not aware of.

\section{Discussion and Conclusion}\label{discussion}

We presented a pragmatic signalling framework that extends Rational Speech Acts (RSA) with temporal belief tracking, user-prior modelling, and finite-horizon planning for dynamic human-AI collaboration. Our systematic evaluation demonstrates that pragmatic planning --- incorporating both user-awareness and lookahead reasoning --- significantly improves communication effectiveness over baseline approaches, particularly in scenarios with challenging environments, diverse user profiles and temporal constraints. This work establishes a foundation for adaptive assistive systems that strategically time and frame alerts to align with individual users' evolving situational awareness, a capability necessary for safety-critical domains where communication breakdowns have severe consequences.

\subsection{Our Contributions}

\changed{Our work makes four key theoretical and methodological contributions to the broader landscape of human-AI collaboration research. First, we \textbf{extend} the \textbf{RSA framework} to \textbf{dynamic settings}. Classical RSA treats utterances as isolated communicative acts, ignoring that temporal dynamics are essential to effective human-AI collaboration \cite{degen2023}. Our d-RSA framework represents the first systematic extension of the pragmatic reasoning capabilities of vanilla RSA across multiple timesteps. By modelling how listener beliefs propagate, d-RSA enables principled reasoning about communication sequences rather than individual messages.}

\changed{Second, to the best of our knowledge, our work is the first to \textbf{jointly optimise} the \textit{what, how,} and \textit{when} of assistive communication while being \textbf{mindful of human cognition}. Most existing approaches in human-AI collaboration optimise either message content (framing) or temporal allocation (scheduling) independently. Information-theoretic methods focus on optimal content selection given fixed timing constraints \cite{gmytrasiewicz2001rational, karten2023emergent}, while scheduling systems optimise temporal delivery assuming predetermined message content \cite{rosenthal2020human, baruwal2024towards}. Recent work points out their interdependence \cite{liu2024effect}: message framing directly impacts user comprehension, while timing determines the cognitive window available for processing.}

\changed{We bridge this gap by recognising communication as a sequential planning problem where every choice has a temporal opportunity cost (see Eq. \ref{eq:cumulative_reward}). Choosing to deliver a detailed, specific utterance may preclude timely alerts about other emerging problems, a trade-off that isolated optimisers cannot capture. By \textbf{jointly reasoning about content, framing, and timing}, our d-RSA framework enables the AI agent to act as a \textbf{cognitive interface}, guiding human attention --- a finite resource --- to where it is most needed in dynamic, high-stakes environments.}

\changed{Third, our systematic evaluation of $800$ diverse simulated trials helps reveal when pragmatic human-aware planning is most beneficial. Our results show it is most advantageous in precisely the scenarios where \textbf{adaptability} matters most: \textbf{high-complexity environments} with multiple simultaneous critical properties, diverse user knowledge states, and tight time constraints. This communication strategy exhibits \textbf{context-appropriate} behaviour consistent with Gricean cooperative principles \cite{dale1995computational} --- using concise alerts when users possess relevant background knowledge (Maxim of Quantity), providing detailed explanations when users lack context, and strategically timing communications to maximise comprehension (Maxim of Manner). Furthermore, our analysis of anticipatory alerting indicates sophisticated reasoning beyond mere reactivity: the system proactively raises awareness of near-critical conditions, positioning users to interpret subsequent brief alerts more effectively; thereby reducing alert frequency and presumably the resulting alarm fatigue \cite{gulcsen2025effect}. This \textbf{adaptive ability} to plan communication sequences hints at a cooperative partnership currently missing from many human-AI systems.}

\changed{Fourth, our work provides a \textbf{principled theoretical grounding for assistive agents}. We contend that effective human-AI systems are built upon explicit, structured models of the task, the world, and the human partner. This ensures that cooperative behaviour is by design, and not a brittle, emergent property common to agents that rely on heuristic or purely data-driven methods \cite{jamakatel2024goal, collins_building_2024}. A growing body of research emphasises aforementioned Gricean norms provide an essential theoretical foundation for collaboration, with norm-adherent agents achieving higher task accuracy and generating more appropriate responses \cite{zhi2024pragmatic, saad2025gricean}. Our choice to build upon the RSA framework stems directly from this insight. As a computational model of Gricean cooperation, RSA's Bayesian recursive structure provides the mechanism for Theory of Mind reasoning, allowing agents to explicitly model the mental states of their communicative partners \cite{degen2023}.}

\changed{This principled grounding offers a crucial advantage in reliability and predictability, particularly when contrasted with the often opaque reasoning of purely neural models and LLMs. While LLMs excel at generating fluent referring expressions \cite{liu2024}, they also exhibit systematic pragmatic violations, such as excessive specificity, and can produce user misunderstandings more frequently than human partners \cite{tang2024, kapur2025}. Our approach ensures adherence to cooperative principles, vital for safety-critical applications. The primary trade-off, of course, is the need for hand-crafted lexicons, a limitation that itself opens up several promising directions for future research.}

\subsection{Future Directions}

\changed{Most immediately, this points toward hybrid approaches -- combining RSA's principled reasoning with the generative flexibility of neural models, reducing manual lexicon construction while maintaining pragmatic consistency \cite{cohngordon2018}.
Next, future work should relax our model's assumptions of perfect knowledge of user awareness and future world states --- idealisations that enable controlled comparison in simulated environments but may not hold in practice. On the user side, our static knowledge model can be extended by incorporating dynamic user attention models, enabling real-time adaptation through behavioural monitoring (e.g., eye-tracking \cite{das2024shifting}). On the environment side, moving beyond the assumption of a perfectly known future requires addressing the fundamental challenges of planning under uncertainty. Monte Carlo tree search \cite{kocsis:szepesvari:ecml-06,keller:helmert:icaps-13} approximations could enable scalable deployment while managing the uncertainty inherent in real-world environments.
Theoretically, extending our binary criticality model to multi-level urgency hierarchies would better reflect real-world operational priorities -- distinguishing, for e.g., between a gradual battery depletion and a sudden rotor malfunction.} 

\changed{To conclude, by integrating principled pragmatic reasoning with temporal planning, this work offers a path toward assistive agents that truly cooperate with human cognition rather than overwhelming it. In an era where human-AI interdependence gradually becomes inescapable, ensuring our AI agents are effective and collaborative is a scientific necessity. Our theoretically grounded approach aims to provide a concrete step in that direction.}

\bibliographystyle{ACM-Reference-Format} 
\bibliography{rsa-plan}

\section{Code and Data Appendix}

Repository: url to be public, upon acceptance

Files required to replicate are also attached as \textit{.zip} along with this Supplementary Material document.

\section{Technical Appendix}

\subsection{Alternative Model Formulations and Design Decisions}

This section documents the alternative formulations we considered for key model components during ideation and development, the rationale for exploring each variant, and why ultimately we selected our current implementations. It is our aim that we make the design process transparent, for it to serve as a stepping stone for future researchers exploring other, more realistic alternatives.

\subsubsection{The Lexicon $M$ and Costs}
Our framework intentionally diverges from the classical RSA approach of incorporating utterance costs as exponential discounts in the speaker's softmax distribution ($S(u|p) \propto \exp\left( \alpha \left( \log P(p| u) - C(u) \right) \right)$) [Note: we assume the rationality parameter $\alpha$ to be 1, for simplicity]. This traditional cost discounting becomes challenging in our dynamic belief-tracking context, where reward values fluctuate greatly across trials and property types (differing belief value ranges) --- determining fixed cost scales relative to rewards becomes unreliable and difficult to calibrate. Instead, we model opportunity costs explicitly as planning constraints on the Pragmatic Speaker. This approach is advantageous not only because it is scale invariant, but it is also interpretable, i.e., resource limitations are explicit rather than buried in cost parameters.

The addition of \textit{`(X)'} (Block constraint for multi-step alerts) also allows for our lexicon $\mathcal{M}$ to reflect graded communicative specificity through fully explicit (\textit{`Drone~2 Battery Low'}) two-feature alerts or single-feature alerts like \textit{`Battery'} (attribute-level, requires inference) compared to generic signals like \textit{`Beep'}, and more importantly, silence \textit{`(...)'} to model non-communication. These help ensure our model remains prima facie plausible as a representation of a listener’s subjective attentional capacities.

\subsubsection{Components of the Model}

Our initial approach combined world-state evolution and user belief updating in a single \textit{complex} transition function $T$ that attempted to optimise user comprehension through multi-step lookahead. This approach conflated environmental dynamics with user or listener cognition, making the system less interpretable and harder to debug while also moving away from an RSA-like structure. The modular decomposition into $AT$, $B$, and $R$ proved more flexible and better aligned with RSA's principled framework; each of which we elaborate next.

\paragraph{Attention Mapping $AT$:}
The attention function $AT$ transforms the $L_1$ distribution into user attention allocation across properties. We explored two primary formulations:

\begin{enumerate}
    \item \textbf{Direct Mapping (Selected):} $at_t(p) = L_1(p \mid u_t, b_{t-1}, StatCr)$. Provides direct correspondence between pragmatic listener interpretation and user attention allocation, preserving full probabilistic information from $L_1$ without additional transformation parameters.
    
    \item \textbf{Binary Attention (Winner-Take-All):} $at_t(p) = \mathbf{1}[p = \arg\max_{p'} L_1(p')]$. A qualitative review of initial trials ($n=12$) showed this was overly restrictive -- users lost all information about secondary properties even when utterances conveyed meaningful ambiguity, leading to overly conservative message selection.
\end{enumerate}

The choice of $AT$ directly influences the granularity of belief updates and reward computation. Future work could investigate how different attention allocation schemes affect model performance across varying cognitive load conditions (e.g., high load $\to$ binary, etc.).

\paragraph{Estimated Belief Update $B$:}
In our main paper, we adopt a linear interpolation approach for belief updates (Equation 4 (main paper)), to balance cognitive plausibility with simplicity with perspicuity. However, we also considered a more psychologically motivated alternative inspired by bounded rationality models \cite{mobius2014managing}, which combines evidence multiplicatively.

\begin{equation*}
\begin{aligned}
& \hspace{-0.6cm} \text{logit}(b_t(p)(v)) = \delta \cdot \text{logit}(b_{t-1}(p)(v)) + \\
& \hspace{1.2cm}
\gamma_{Cr} \cdot Cr_t(p) \cdot \text{logit}(P(s_t(p) = v))
\end{aligned}
\label{eq:beliefs}
\end{equation*}

where $\delta$ controls prior belief influence, $\gamma_{Cr}$ modulates learning rates based on criticality; and while this formulation captures realistic phenomena like memory decay and attention-modulated learning --- it introduces multiple hyper-parameters that are difficult to estimate and tune during planning. For the purposes of evaluating our framework, the linear interpolation model provides sufficient psychological realism while maintaining interpretability, and this alternative remains as a valuable component for simulating dynamic user beliefs online for evaluating our full model in the future.

\paragraph{Reward Function $R$:}
The reward function determines which utterances best serve task objectives. For our task, we explored the following formulations during development:

\begin{enumerate}
    \item \textbf{Critical-Only Reward (Selected):} $R_t(u_t) = \sum_{p} b_t^{u_t}(p)(s_t(p)) \cdot Cr_t(p)$. Focuses on how well utterances align listeners' beliefs to the true world state, especially for critical properties.
    
    \item \textbf{Belief Delta Reward:} $R_t(u_t) = \sum_{p} [b_t^{u_t} - b_{t-1}](p)(s_t(p)) \cdot Cr_t(p)$. Qualitative review of initial trials ($n=12$) revealed gaming behaviour where the system would alert users about near-critical properties before they became critical to maximise subsequent reward gains due to larger \textit{shifts}. This violates the core conversational maxim of truthfulness, and equipping systems in safety-critical contexts could undermine operator (human user) trust.
    
    \item \textbf{Additive Weight Rewards:} $R_t(u_t) = \sum_p b_t^{u_t}(p)(s_t(p)) \cdot [Cr_t(p) + \omega]$. Designed to maintain situational awareness during low-criticality periods. However, small weights caused generic, unambiguous messages (e.g., \textit{`Beep'}) to achieve higher rewards by distributing probability mass broadly across all properties, diluting focus on urgent hazards. As expected, increasing the $\omega$ proved more suitable, but it achieved similar results to the Critical-Only Rewards without the extra parameter to be determined during run. 
\end{enumerate}

These explorations also confirmed that attention is indeed a scarce resource in time-critical domains, justifying our focus on critical properties exclusively. 

\subsection{Experimental Protocol}

This section details the experimental setup used to generate trails and then evaluate our model's belief tracking and pragmatic planning capabilities across a broad range of scenarios and user profiles.

\subsubsection{User Belief Modelling, in detail.}
The $L_1$ begins with an initial belief $b_{t=0}(p, v)$, a probability distribution over each property $p$ and its possible values $v$. We vary two aspects of these initial beliefs to simulate diverse users with different levels of situational awareness:
\begin{itemize}
    \item Critical Property $Cr(p)$ Awareness: We vary the probability that a user holds the accurate initial beliefs about current and soon-to-be $Cr(p)$ across $10\%, \; 25\%, \; 50\%, \; 80\%$ \emph{i.e.} ranging from uninformed to situationally well-aware users.
    \item General Awareness: For all other properties ($p \notin Cr(p)$), we generate noisy belief distributions centred around the true value $s_{t=1}(p)$. Users are classified as having \textit{low} or \textit{high} general awareness. Low-awareness users have a belief means $\mu$ drawn from a wide range around $s_{t=1}(p)$ and high standard deviations $\sigma$ (both imprecise and uncertain). High-awareness users have $\mu$ closely centered on $s_{t=1}(p)$, with narrow $\sigma$ (more precise and confident).
\end{itemize}

\subsubsection{Environment Parameters, summarised.}
\begin{itemize}
    \item Environment: $4$ drones $\times$ $6$ attributes $=$ $24$ total properties
    \item Property types: Battery, Wind-Speed, Rotor, Altitude, No-Fly-Zone, Distance
    \item Horizon: $7$ timesteps
    \item Critical properties per scenario: $\{2, 3, 4\}$ (equally distributed)
    \item Temporal dispersion levels:$\{0, 1, 2, 3\}$ timestep intervals
    \item First critical onset: uniform random $\in [1, 3]$.
\end{itemize}

\subsubsection{Models and their Evaluation}
Table~\ref{tab:model_reason} summarises the reasoning capabilities of each model variant.

\begin{table}[h!]
\centering
\caption{Models and their Reasoning Capabilities}
\label{tab:model_reason}
\setlength{\tabcolsep}{3pt}
\begin{tabular}{lccc}
\toprule
\textbf{Models} & \textbf{Planning} & \textbf{User Priors} \\
\midrule
Baseline (d-RSA)         & \ding{55}                  & \ding{55}               \\
d-RSA + Priors           & \ding{55}                  & \ding{51}               \\
d-RSA + Planning          & \ding{51}                  & \ding{55}               \\
Full (d-RSA + Priors + Planning)            & \ding{51}                  & \ding{51}               \\

\bottomrule
\end{tabular}
\end{table}

800 unique scenarios multiplied by 4 model variants yields 3,200 experimental runs. Each run records full belief trajectories along with all legal utterance sequences for $1 \leq t \leq H$.

We analyse results using linear mixed-effect regression models implemented in \texttt{lme4} (version 1.1.35.5, in R version 4.4.3) with random intercepts for scenario-level variance. Statistical significance is reported using two-tailed $p$-values with 95\% confidence intervals computed via profile likelihood.

\subsubsection{Infrastructure and Parameters}

Experiments were conducted using HTCondor, with the task being parallelised across $\simeq 40$ individual job instances at a time, allocated to a pool of Linux machines. Each job was allocated a minimum of $30$ CPU cores and $640$ GB RAM. We used Python  3.9.21 along with standard scientific computing libraries. No GPU acceleration was employed.




\end{document}